\documentclass[11pt,a4paper]{article}
\usepackage{times,latexsym}
\usepackage{url}
\usepackage[T1]{fontenc}
\usepackage{tabularx}
\usepackage{graphicx}
\usepackage[utf8]{inputenc}
\usepackage{microtype}

\usepackage{emnlp2021}

\usepackage{xspace,mfirstuc,tabulary}

\newif\iftaclinstructions
\taclinstructionsfalse % AUTHORS: do NOT set this to true
\iftaclinstructions

\newcommand{\instr}
\fi

\title{Detecting Multidimensional Political Incivility on Social Media}

\author{Sagi Pendzel \\
Computer Science Dep. \\
University of Haifa 
  \And
  Nir Lotan \\
Information Systems Dep. \\
University of Haifa 
  \\\AND
  Alon Zoizner  \\
Communication Dep. \\
University of Haifa 
  \\\And
  Einat Minkov \\
Information Systems Dep. \\
University of Haifa 
\\}
\date{}

\begin{document}
\maketitle
\begin{abstract}
The rise of social media has been argued to intensify uncivil and hostile online political discourse.  Yet, to date, there is a lack of clarity on what incivility means in the political sphere. In this work, we utilize a multidimensional perspective of political incivility, developed in the fields of political science and communication, that differentiates between impoliteness and political intolerance. We present state-of-the-art incivility detection results using a large dataset of 13K political tweets, collected and annotated per this distinction. Applying political incivility detection at large-scale, we observe that political incivility demonstrates a highly skewed distribution over users, and examine social factors that correlate with incivility at subpopulation and user-level. Finally, we propose an approach for modeling social context information about the tweet author alongside the tweet content, showing that this leads to improved performance on the task of political incivility detection. We believe that this latter result holds promise for socially-informed text processing in general.\footnote{The dataset is available at \href{https://huggingface.co/incivility-UOH}{Hugging Face}, and additional supplementary materials can be accessed 
\href{https://github.com/sagipend/detecting-multidimensional-political-incivility-on-social-media}{here}.}
\iffalse\footnote{Dataset and other supplementary materials can be accessed \href{https://anonymous.4open.science/r/detecting-multidimensional-political-incivility-on-social-media-72DD}{here}.}\fi
\end{abstract}

\section{Introduction}

A civil discourse between political groups is considered a fundamental condition for a thriving and healthy democracy~\cite{gutmann2009}. Sadly, the rise of social media has been argued to intensify disrespectful and hostile online political discourse~\cite{coe2014,frimer2023}. According to researchers, there are multiple negative consequences of this phenomenon to democracy: it fosters polarization between rival political groups, decreases trust in political institutions, and may disengage citizens from being politically involved~\cite{muddiman2020,skytte2021,van2022effects}.

Considering these concerns, scholars have attempted to quantify uncivil political discourse in discussion groups and social media platforms~\cite{elsherief2018,davidsonWS2020,theocharis20,frimer2023}. These efforts offer however a coarse definition of incivility. Political communication researchers rather view political incivility as a multidimensional concept~\cite{muddiman2017,rossini}. The first dimension is {\it personal-level incivility (impoliteness)},  pertaining to a violation of interpersonal norms. Impolite speech may contain foul language, harsh tone, name-calling, vulgarity, and aspersion towards other discussion partners or their ideas (e.g., ``are you really so stupid that you would defund this program?''). The second dimension of {\it public-level incivility (intolerance)} refers to violations of norms related to the democratic process, such as pluralism and deliberation. It refers to exclusionary speech, silencing social and political groups and denying their rights~\cite{rossini} (e.g., ``Hillary and the dems ARE enemies, foreign AND domestic''). Considering these separate dimensions is crucial when detecting incivility on digital platforms since they carry different democratic implications. In fact, political impoliteness may sometimes lead to  positive outcomes, such as increasing citizens' interest in heated debates and opinion justification~\cite{Papacharissi,rossini}. 

This work makes several contributions to the study of political incivility on social networks. First, we address political incivility detection at fine-grained resolution. We constructed a dataset of 13K political tweets from the U.S. context for this purpose, which we labeled via crowdsourcing. The data collection process involved diverse sampling strategies, aiming at capturing sufficient examples of both incivility types while avoiding lexical biases. We make this resource available to the research community. We then finetuned state-of-the-art transformer-based language models on the task of multi-label incivility detection. Due to the size and diversity of our dataset, we achieve state-of-the-art results both within- and across-datasets. Our experiments illustrate the differences and performance gaps in identifying impolite speech, which is typically explicit, and political intolerance, which often requires social and semantic understanding.

A second contribution is our focus not only on individual tweets to study political incivility but also on the user level. Applying political incivility detection at large scale, we examine the prevalence of incivility among more than 200K random American users who posted political content on Twitter. Shifting the focus to the user level allows us to answer important research questions: (i) Are there differences in incivility levels between subpopulations of interest--Democrats vs. Republicans, or across states? (ii) Are some individual users more inclined than others to using impolite and intolerant language in political discussions on social media? (iii) Can relevant user representations be effectively modeled as context, so as to perform author-informed detection of political incivility? Our investigation of these questions leads to a formulation of social text processing, where textual contents and social information about the text author (based on his social network) are modeled jointly in identifying political incivility. We show that such an approach can lead to substantial performance gains, both in terms of precision and recall.

\section{Related work}
\label{sec:related}
	
To the best of our knowledge, this work is the first to implement a multidimensional perspective for political incivility detection and evaluation at scale. \iffalse 
to detect, characterize and quantify online political incivility at scale.\fi Notably, public-level political incivility (intolerance) is a broad concept. While there exists ample related research on the detection of {\it hate speech}, an exclusionary speech against social minorities~\cite{fortuna18}, relatively few research works sought to generally detect, characterize and quantify uncivil online discourse in the context of a politically polarized climate. Several previous works aimed at detecting political incivility in online platforms, however these works have either considered impoliteness and intolerance as a unified concept~\cite{Theocharis2016,theocharis20}, or focused only on one of these dimensions~\cite{davidsonWS2020,elsherief2018,shvets2021}. This conceptual and methodological fuzziness ignores the different democratic outcomes of each of these dimensions. Whereas insults and foul language (impoliteness) may be considered acceptable in polarized environments and  heated political debates~\cite{rains2017}, expressions that refuse to recognize the legitimacy of a rival group or consider it morally inferior (intolerance) are far less acceptable~\cite{prooijen2019}.

Our exploration of political incivility at user level, and the modeling of text alongside the social encoding of its author, form another important contribution of this work.

\section{A dataset of fine-grained political incivility}

This section describes our steps of data collection and annotation in constructing a labeled dataset of multidimensional political incivility. \iffalse In this section, we describe our data sampling procedure intended at achieving high diversity, while reducing lexical biases. We then discuss the details of data labeling via crowd sourcing, the observed inter-annotator rates, and the resulting dataset statistics.\fi

\subsection{Data sampling strategy}

Even though incivility is not rare, the inspection of random tweets would yield a low ratio of relevant examples at high annotation cost. We exploit multiple network- and content-based cues, aiming to obtain a diverse sample of relevant tweets while avoiding lexical and other biases~\cite{wiegandNAACL19}.

{\it Obtaining political tweets.} 
\label{par:political}
First, we retrieved a large pool of tweets which we expected to include fervent political language. Concretely, we referred to several lists of social media accounts in the political domain that are disputable or biased, including: accounts that are known to distribute fake news~\cite{grinberg2019fake}, the accounts of members of the \iffalse 117th\fi U.S. Congress who are considered ideologically extreme~\cite{voteview},\footnote{https://voteview.com/data} and news accounts that are considered to be politically biased to a large extent~\cite{wojcieszak2021no}. \iffalse Each account was assessed by political bias measured in terms of dw-score(?).\fi We selected the 20 most biased accounts per category, of either conservative or liberal orientation, based on bias scores provided by those sources.\footnote{The fake news category includes only 9 accounts of each orientation.} \iffalse , maintaining a balance between Republican and Democratic orientation.\fi We then identified users in our pool who follow at least two of the specified biased accounts, maintaining a balance between users of conservative and liberal orientation. Retrieving 
the (200) most recent tweets posted by the sampled users, using Twitter API as of December 2021, yielded 885K tweets authored by 15.8K users. Finally, applying a dedicated classifier (Sec.~\ref{sec:political}), we identified 82K of those tweets as political. Annotating 300 random tweets of this pool by a graduate student of Communication indicated on precision of 0.91 (i.e., 273 of the 300 tweets were confirmed to be political). 

{\it Sampling tweets for annotation.} Aiming to further focus on political tweets that were likely demonstrate incivility, we again applied several sampling guidelines. The selected tweets were then subject to manual annotation by crowd workers (Sec.~\ref{sec:crowdsource}). First, similar to previous works~\cite{theocharis20,hedeEACL21}, we utilized the pretrained Jigsaw Perspective tool\footnote{https://www.perspectiveapi.com/} to identify toxic language. \iffalse such as severe toxicity, insult, profanity, identity attack, threat, inflammatory language, attack on author, and obscenity.\fi Specifically, we considered tweets that received relatively high scores on the categories of `abusive language and slurs', `inflammatory comments' and `attacks on the author'. \iffalse Our objective was to encompass a diverse range of tweets, rather than just those with a high confidence of being toxic, to avoid limiting ourselves to a narrow group of users who frequently use foul language.  The tweets that scored higher than those thresholds were selected. \fi Roughly 1.9K tweets were sampled in this fashion, where the human annotators labeled 43.3\% and 9.9\% of them as impolite and intolerant, respectively. In addition, following the insights inferred by Ribiero {\it et al}~\citeyearpar{ribeiro2018characterizing} with respect to hateful tweets, \iffalse by which hateful user accounts tend to be created later in time compared with other accounts, possibly because of being suspended over time, and that such accounts tend to post content more frequently than average. We therefore\fi we favored the sampling of tweets by user accounts that were new, being created up to two months prior to sampling date, or highly active, having posted more than one tweet a day on average since the account creation date. Annotating 2.0K tweets selected based on these criteria yielded proportions of 25.9\% and 7.5\% of impolite and intolerant tweets, respectively. \iffalse  3) The tweet has at least 50 words or at least a third of the words are in capital letters.\fi \iffalse We applied our political classifier to 11.4M tweets posted by 1.3M users from the Infomedia corpus, identifying 85.8K tweets as political.  We then used the Perspective API to filter tweets that were likely to be uncivil. We defined political users as those who authored at least one political tweet, and uncivil users as those with at least one uncivil political tweet. We then attempted to identify accounts that uncivil users tend to follow using the PMI measure. We then identified users who follow the largest number of of accounts that characterise uncivil users, and extracted their tweets.\fi Finally, we sampled 4K tweets from the pool of political tweets uniformly at random, where this yielded lower ratios of relevant labeled examples: 12.9\% impolite and 3.2\% intolerant tweets.\footnote{More precisely, we attempted sampling 2K tweets of those based on additional network cues rather than randomly, however this resulted with similar identified incivility rates.} 

{\it Active sampling.} Overall, the annotation of 7.9K political tweets sampled as described above yielded 2.3K examples labeled as impoliteness (28.9\%) and 0.8K examples labeled as political intolerance (9.8\%) (this includes 2.8\% of the tweets that were labeled as both categories). In order to obtain more examples of political intolerance, we employed a classifier of intolerance detection trained using these labeled examples~\cite{tong01}. In several consequent batches, we sampled 5.2K tweets which the classifier identified as intolerant. Overall, the ratio of identified impoliteness in those tweets was similar (22.5\%), where the observed ratio of intolerance has tripled (29.5\%) (2.1\%  labeled as both categories). The final dataset statistics are detailed in Table~\ref{tab:stats}. Importantly, we allocated all of the labeled examples obtained via active sampling to the training set in our main classification experiments in order to avoid evaluation bias. 

\subsection{Identifying political tweets} 
\label{sec:political}

As we study incivility in political contexts, it is first required to identify topical relevance. Topic detection is a well-studied task, for which excellent performance can be achieved given a sufficient number of labeled examples using models such as BERT~\cite{bert}. To obtain labeled examples, we referred to an existing dataset of political tweets collected by Barberá {\it et al.}~\citeyearpar{barbera2015tweeting}, randomly sampling 12.5K tweets across different political topics,\footnote{Topic titles: `2012 presidential campaign', `2013 government shutdown', `Budget' and `Marriage equality', `the Boston Marathon bombing', `Newtown school shooting,' and `use of chemical weapons during the Syrian civil war'.} \iffalse Non-political topics in the dataset include the 2014 Super Bowl, Winter Olympics, and Academy Awards.\fi and further sampled 3.5K political posts from another public dataset of political social media posts.\footnote{www.kaggle.com/datasets/crowdflower/political-social-media-posts}. We considered random Twitter tweets by U.S. users as counter non-political examples, constructing a balanced dataset of 32K political and non-political tweets overall. While this labeling strategy is noisy, contrasting topical tweets with random examples should support effective learning, as confirmed by our results. 

\iffalse Prior to classification, we applied several pre-processing steps to the text. Retweets, HTML tags, embedded URLs, hashtags, mentions, and non-English characters, were removed from the text. In addition, we processed up to 512 tokens per tweets, truncating any longer tweets accordingly. In learning, we split the dataset into stratified train (70\%), validation (10\%) and and test (20\%) sets.\fi We fine-tuned a BERT-base uncased model using its public implementation and standard training practices, minimizing the Cross-Entropy loss function. \iffalse applying an AdamW optimizer with a learning rate of 2e-5, and a batch size of 16 for 3 epochs.  on Google Colaboratory tool with an NVIDIA Tesla P100 GPU and 16G RAM as the implementation environment.\fi  Evaluation of political tweet detection on held-out examples (20\%) indicated on high precision and recall scores of 0.97. Aiming at maintaining high precision in detecting political tweets, also in data shift conditions, we set a high threshold (0.96) over the classifier's confidence. As reported before, the precision of the classifier was assessed at 0.91 on our pool of candidate tweets.  

\iffalse
\begin{table}[t]
%\centring
\begin{footnotesize}
\begin{tabular}{lrrc}
Source & uncivil & uploaded  & \% uncivil \\
\hline 
liberal fake news  & 345 & 762 &	45.28   \\
conservative news & 1182 & 2645 &	44.69     \\
republican members  & 2122 & 4801 & 44.20   \\
conservative fake news  & 321 & 778  &	41.26    \\
democratic members  & 1483 & 3804 &	38.99   \\
liberal news & 94 & 334 &	28.14   \\
\hline
\end{tabular}
\end{footnotesize}
\caption{Uncivil labeled tweets distribution among followers of accounts who are considered ideologically extreme from both sides - liberals and conservatives (as described in section 4.2) 
** tweet considered as uncivil if it was labeled as impoliteness or intolerance or both.  }
\label{tab:twitter_accounts_mturk_statistics}
\end{table}
\fi

\subsection{Crowdsource labeling}
\label{sec:crowdsource}

\begin{figure}[t]
% \hline
\iffalse\includegraphics[width=0.5\textwidth]{figures/interface_small.png}\fi
\includegraphics[width=0.48\textwidth]{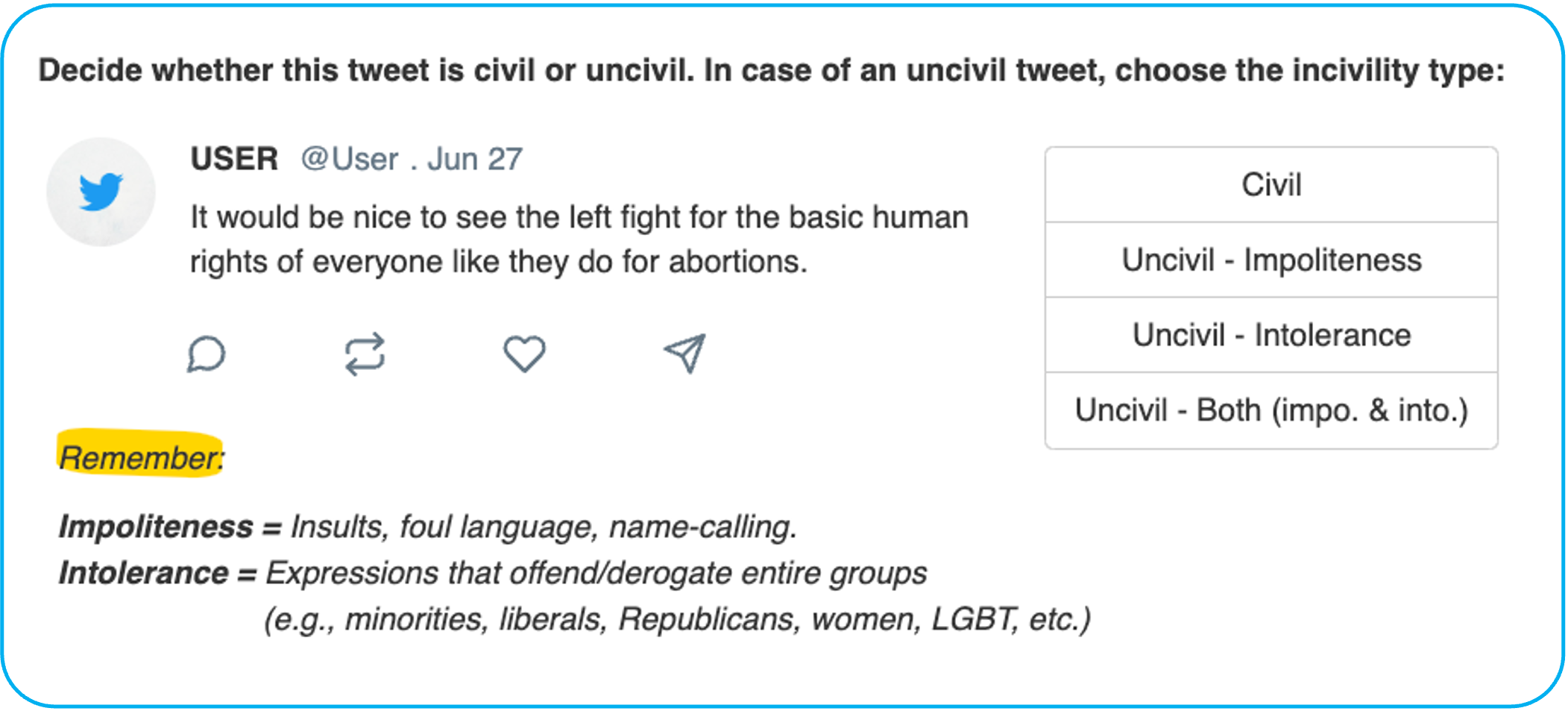}
% \hline
\caption{Annotator interface: the workers were asked to label tweets as impolite, intolerant, neither or both.}
\label{fig:interface}
\end{figure}

We employed non-expert workers on the Amazon Mechanical Turk platform\footnote{www.mturk.com/} to obtain human judgements regarding political incivility. Given each selected tweet, several independent workers were asked to determine whether it was impolite, intolerant, both, or neither.\footnote{The category of `both' was specified in order to raise annotator awareness of this possibility.} Table~\ref{tab:examples} includes examples which we presented to the workers of each class. These examples were accompanied with a codebook containing explanations regarding the guidelines for annotating the tweets. Figure~\ref{fig:interface} shows the annotation interface that workers were presented to workers for labeling the tweets.

\begin{table*}[t]
%\centring
\begin{footnotesize}
\begin{tabularx}{\textwidth}{X}
\hline
IMPOLITE:  ``All hell has broken loose under the leadership of the senile old man. I don’t believe a damn word from this dumb son of a bitches.''; ``That's what they are protesting, you rank imbecile. People like you need a damn good kicking.'' \\
\hline
INTOLERANT: ``Hillary and the dems ARE enemies, foreign AND domestic''; ``If you agree with democrats in congress, you are an anti-American commie'' \\
\hline
NEUTRAL:  ``How long do Republicans believe you can keep pushing this line? You never intended to secure the border''; ``There are 400,000,000 guns in the United States, you’re going to have to stop the criminals not the guns''\\
\hline
\end{tabularx}
\end{footnotesize}
\caption{Example tweets per class}
\label{tab:examples}
\end{table*}

Crucially, the task of assessing political incivility in general, and differentiating between impoliteness and intolerance in particular, involves fine semantics and critical thinking.  We therefore took several measures to assure high quality of the annotations. First, we restricted the task to highly qualified workers (who had previously completed at least 100 HITs with approval rate higher than 98\%). We also required the assigned workers be residents of the U.S., to assure that they were fluent in English and familiar with U.S. politics. Relevant candidate workers were further asked to undergo a training and qualification phase. Each candidate worker was asked to label six carefully selected tweets, where in case of a mistake, they received feedback with an explanation about the correct label. Whoever labeled a majority of the tweets correctly got qualified to work on our task. \iffalse enter our annotation work's HITs. \fi Finally, we included control questions (two out of 15 tweets in each micro-task, also referred to as a HIT) that we expected the workers to do well on. In case that the worker failed to label the control tweets correctly, \iffalse of a mistake on a control tweet or in case of a worker drop below 20\% of correct answers on the test tweets,\fi we discarded the annotations, and banned those workers from further working on our task. We paid the workers an hourly fee of 17.5 USD on average, which exceeds the U.S. minimum wage standards, where fair pay is known to positively affect annotation quality~\cite{ye2017does}.  \iffalse \textcolor{red}{how much exactly? - On average, it takes around 3.5 minutes to complete one HIT, and we pay workers 1 dollar for each HIT completed. So, approximately 17.5\$ per hour.}  Overall, 550 workers have taken our qualification test. Over the course of data labeling of multiple HITs, a majority of the workers were eliminated, resulting in a final cohort of 125 workers.\fi Overall, our final cohort included 125 workers, who annotated up to 2,000 tweets per week, over a period of 3 months. \iffalse who received tweets in batches that included 200-300 HITs, each containing 15 tweets. so they won’t able to annotate more than 2000 tweets per week. The whole annotation process lasted 3 months.\fi

\subsection{Dataset statistics}

Each tweet was labeled by 3-5 annotators, where we assigned the final labels using majority voting. \iffalse Accordingly, we obtained up to five annotations in order to break ties for tweets for which no consensus has been reached.  where we discarded examples where this was not possible (25.8\% of the total).\fi
Overall, our dataset includes 13.1K labeled tweets. As detailed in Table~\ref{tab:stats}, a large proportion of the labeled examples (42.3\%) corresponds to political incivility, including 3.6k tweets labeled as impolite, and 2.3K as intolerant. In comparison, existing related datasets are smaller, use binary annotations, and include substantially fewer incivility examples.

\iffalse Overall, the dataset includes 7580 civil and 8365 uncivil tweets. However, when we consider the finer-grained distinction between incivility types, namely impoliteness vs. intolerance, we failed to resolve the label via majority voting in 2.3K cases, thus multi-label dataset size is reduced to 13.1K examples. \fi

\iffalse
\begin{table}[t]
%\centring
\begin{footnotesize}
\begin{tabular}{lrrcc}
 & \multicolumn{2}{c}{Labeled tweets} & \multicolumn{2}{c}{No. of labels} \\
Label & Number & Ratio & Average & Std \\
\hline 
Civil & 7577 & 57.7\% & 2.98 & 0.54 \\
Impolite & 3234 & 24.6\% & 3.07 & 0.62 \\
Intolerant & 1977 & 15.1\% & 3.21 & 0.64 \\
Both & 336 & 2.6\% & 3.01 & 0.47  \\
\hline
All &	13124 & 100.0\% & & \\
\end{tabular}
\end{footnotesize}
\caption{Dataset statistics: label distribution.}
\label{tab:stats0}
% \blfootnote{Labels distribution}
\end{table}
\fi
\iffalse
\begin{table}[t]
%\centring
\begin{footnotesize}
\begin{tabular}{lrrcc}
 & \multicolumn{2}{c}{No. of coders} \\
Label & Average & Std \\
\hline 
Civil & 2.98 & 0.54 \\
Impolite & 3.07 & 0.62 \\
Intolerant & 3.21 & 0.64 \\
Both & 3.01 & 0.47  \\
\hline
\end{tabular}
\end{footnotesize}
\caption{Dataset statistics: label distribution}
\label{tab:stats0}
% \blfootnote{Labels distribution}
\end{table}
\fi

\begin{table}[t]
%\centring
\begin{footnotesize}
\begin{tabular}{lrrc}
Dataset & Size & Uncivil & Impol./Intol. \\
\hline 
Ours & 13.1K & 42.3\% & 27.2 / 17.7\% \\
\hline
\citeauthor{davidsonWS2020} & 1.0K & 10.4\% & - \\
\citeauthor{lineoffire} (USA) & 5,0K & 15.4\% & - \\
\citeauthor{lineoffire} (CAN) & 5.0K & 10.6\% & - \\
\citeauthor{theocharis20} & 4.0K & 17.4\% & - \\
\hline
\end{tabular}
\end{footnotesize}
\caption{Dataset statistics: ours vs. other datasets}
\label{tab:stats}
% \blfootnote{Labels distribution}
\end{table}

\iffalse
\begin{table}[t]
%\centring
\begin{footnotesize}
\begin{tabular}{lcccc}
 & Precision & Recall & F1  \\
\hline 
Impolite & 0.71 (0.92) & 0.76 (0.79) & 0.74 (0.85) \\
Intolerant & 0.76 (0.80) &	0.74 (0.70) & 0.75 (0.74)  \\
Civil & 0.93 (0.95) &	0.69 (0.79) & 0.79 (0.86)  \\
\hline
%Macro-avg & 0.800 & 0.730 & 0.758 \\
\end{tabular}
\end{footnotesize}
\caption{Evaluation of the labels by crowd workers against gold labels assigned by a domain expert to a sample of 300 random tweets from our dataset.}
\label{tab:ITA2}
\end{table}
\fi

%\textbf{Inter-Annotator Agreement.} 
\iffalse We first evaluate inter-annotator agreement with respect to the binary distinction the text being perceived as civil vs. uncivil. And, considering that a given tweet may be either impolite or intolerant, or both, we evaluated annotator agreement with respect to each of these two labels. \fi

\iffalse In our annotation setup, tweets were labeled using several annotators drawn randomly from a large pool of human workers.\fi To measure inter-annotator agreement, we consider the labels assigned to individual tweets by random worker pairs. \iffalse Since there are thousands of tweets labeled in our dataset, this gives sufficient and credible statistics.\fi \iffalse In general, a score in the range of 0.4-0.6 denotes moderate agreement, and a score between 0.6-0.8 denotes substantial agreement.\fi Our assessment indicated on Fleiss' kappa agreement score of 0.57, reflecting moderate--nearing substantial--agreement, in judging the coarse notion of incivility. Considering our fine-grained annotation scheme, we obtained a substantial agreement score of 0.63 on the category of impoliteness, and moderate score of 0.54 on political intolerance in distinguishing between the target class and the other labels. This suggests that intolerance is more subjective and subtle compared to impoliteness. \iffalse Accordingly, a larger number of annotations (3.21) was required on average to assign a label of intolerance compared with the other classes (3.07 or less). \fi \iffalse 
Table~\ref{tab:stats} details the average number of annotations obtained per example for each of these classes, showing similar trends. A larger number of annotations was required in order to break ties. We observe that more annotations were required on average for the intolerance category (3.21) compared with the other classes. \fi 

We further assessed the quality of the crowdsourced labels against the judgement of a domain expert, who is one of the authors, per 300 random tweets drawn from our dataset. \iffalse, including equal amounts of tweets labeled as civil, impolite and intolerant. Gold labeled: 112 impoliteness, 108 intolerance and 135 civil examples.\fi Assessing the workers' performance against the expert's labels in classification terms~\cite{snowEMNLP08} yielded F1 scores of 0.74 and 0.75 on impolite and intolerant speech, respectively. Considering only the subset of the examples on which the workers showed high agreement (a majority of more than 70\%) resulted in substantially higher annotator F1 score of 0.85 on the impoliteness category. Yet, annotator performance on the intolerance class remained similar (F1 of 0.74). Again, this suggests that the notion of political intolerance is more subtle compared with impoliteness. In general, while political incivility may be perceived differently depending on the background and beliefs of the reader~\cite{opreaACL20}, it is unrealistic to expect that a machine learning approach would outperform human judgement. 

\iffalse 
\begin{table}[t]
%\centring
\begin{footnotesize}
\begin{tabular}{lcccc}
 & precision & recall & f1-score  \\
\hline 
Civil & 0.93 & 0.689 &	0.791   \\
Uncivil & 0.794 &	0.959 &	0.869  \\
\hline
macro-avg & 0.862 &0.824 &0.83
\end{tabular}
\end{footnotesize}
\caption{ITA tweet-level statistics3. Binary:Gold vs. Coders }
\label{tab:ITA_statistics3}
\end{table}

\begin{table}[t]
%\centring
\begin{footnotesize}
\begin{tabular}{lcccc}
 & precision & recall & f1-score  \\
\hline 
Civil & 0.95 & 0.792 &	0.864   \\
Uncivil & 0.839 &	0.963 &	0.897  \\
\hline
macro-avg & 0.894 &0.877 &0.88
\end{tabular}
\end{footnotesize}
\caption{ITA tweet-level statistics2. Binary: Gold vs. Coders. Confidence>0.7. (\#tweets: 81 civil, 51 uncivil) }
\label{tab:ITA_statistics3_high_conf}
\end{table}
\fi

\section{Multidimensional incivility detection}
\label{sec:experiments}

Next, we train and evaluate the extent to which neural models can detect political incivility as perceived by humans. We perform multi-label classification, detecting impoliteness and intolerance as orthogonal dimensions, as well as experiment with coarse prediction of political incivility. %similar to previous works.

\subsection{Experimental setup}

We consider the popular
transformer-based pre-trained language models of BERT~\cite{bert}, RoBERTa~\cite{roberta} and DeBERTa~\cite{deberta}. The latter models have been trained on significantly more text data compared to BERT, and introduced enhancements to its training procedure, cost function, and word attention mechanism. We found that the larger architectures of these models yielded minor improvements, and therefore report our results using the  base configurations of BERT and RoBERTa models, which include 110M and 125M parameters, and DeBERTa-v3 which is a slightly larger model, including 140M parameters. \iffalse Rather than using masked language modeling (MLM), DeBERTa-V3 uses replaced token detection (RTD), which helps to learn more diverse representations. We used DeBERTa-V3 which has been pretrained using similar data to RoBERTa, but \fi 
In addition, we experiment with specialized language models, including HateBERT, a BERT model that has been re-trained for abusive language detection using a large-scale corpus of offensive, abusive, and hateful Reddit comments~\cite{hatebert}, and HateXplain, a model of BERT that has been finetuned on the classification of hateful and offensive Twitter and Gab posts~\cite{mathew2020hatexplain}. All models were applied using their public implementation.\footnote{https://huggingface.co/}
In all cases, we finetune the models using our labeled examples~\cite{bert}.
We split our dataset into fixed stratified train (70\%), validation (10\%) and test (20\%) sets, optimizing the parameters of each model on the validation examples. \iffalse Experiments using a weighted binary cross-entropy loss function. We also attempted repeating the learning process using the joint training and validation portions of the dataset (80\%),\fi \iffalse We also performed cross-validation experiments, which yielded similar results, and are therefore omitted. We also experimented with the large variants of the BERT and RoBERTa models, but these models did not achieve significant improvement compared with our best performing classifier. Additional experimental details are included in the appendix.\fi
Considering the class imbalance, we found it beneficial to employ a weighted cross-entropy loss function, setting example weights according to inverse class frequency, so as to increase the penalty on classification errors on the target minority class. 

\iffalse The annotators were asked to select the target communities mentioned in the post. Subsequently, the annotators are asked to highlight parts of the text that could justify their classification decision. ... 
Up to 10 epochs but with an early stopping trigger (most of the experiments were with 3 epochs) and with the same implementation environment (except RoBERTa-large was with batch size=16 and maximum input length for RoBERTa: 80, duo to memory issues). \fi 

\subsection{Classification results}

\begin{table*}[t]
\begin{footnotesize}
    \centering
    \begin{tabular}{l|cccc|cccc|c}
        & \multicolumn{4}{c}{Impolite} & \multicolumn{4}{c}{Intolerant} &  \\
Classifier	&	AUC	&	P	&	R	&	F1	&	AUC	&	P	&	R	&	F1	&	Macro-F1 \\
\hline
BERT	&	0.857	&	0.635	&	0.713	&	0.671	&	0.848	&	0.530	&	0.644	&	0.581	&	0.626 \\
RoBERTa	&	\textbf{0.874}	&	0.642	&	0.744	&	0.689	&	\textbf{0.859}	&	0.501	&	0.728	&	\textbf{0.593}	&	0.641 \\
DeBERTa	&	0.861	&	0.687	&	0.707	&	\textbf{0.697}	&	0.845	&	0.558	&	0.626	&	0.590	&	\textbf{0.643} \\
HateBert	&	0.865	&	0.701	&	0.661	&	0.680	&	0.835	&	0.515	&	0.639	&	0.571	&	0.625 \\
HateXplain	&	0.820	&	0.567	&	0.688	&	0.622	&	0.756	&	0.374	&	0.537	&	0.441	&	0.531 \\
        \hline
    \end{tabular}
    \caption{Multi-label classifiers performance on our test set}
\label{tab:multi}
\end{footnotesize}
\end{table*}

\iffalse
\begin{table*}[t]
\begin{footnotesize}
    \centering
    \begin{tabular}{l|c|ccc|ccc}
        & & \multicolumn{3}{c}{Civil} & \multicolumn{3}{c}{Uncivil}  \\
Classifier	&	AUC	&	P	&	R	&	F1	&	P	&	R	&	F1	\\
\hline
BERT	&	0.849	&	0.788	&	0.834	&	0.81	&		0.752	&	0.692	&	0.721	\\
RoBERTa	&	0.864	&	0.798	&	0.841	&	\textbf{0.819}	&		0.765	&	0.707	&	0.735	\\
DeBERTa	&	\textbf{0.865}	&	0.812	&	0.824	&	0.818	&		0.754	&	0.739	&	\textbf{0.746}	\\
HateBert	&	0.857	&	0.802	&	0.830	&	0.816		&	0.755	&	0.719	&	0.737	\\
HateXplain	&	0.811	&	0.722	&	0.886	&	0.796		&	0.773	&	0.532	&	0.630	\\
Theocharis	&	0.699	&	0.730	&	0.810	&	0.768		&	0.694	&	0.589	&	0.637	\\
Davidson	& 0.532 &	0.595	&	0.993	&	0.744	&	0.889		&	0.072	&	0.134 \\
\hline
    \end{tabular}
    \caption{Binary classification performance on test set}
\label{tab:binary0}
\end{footnotesize}
\end{table*}
\fi

\begin{table}[t]
\begin{footnotesize}
    \centering
    \begin{tabular}{lccccc}
%        & \multicolumn{3}{c}{Uncivil}  \\
Classifier	&	P	&	R	&	F1	& Mac.-F1 & AUC \\
\hline
BERT	&	0.752	&	0.692	&	0.721	& 0.766 & 0.849\\
RoBERTa	&	0.765	&	0.707	&	0.735	& 0.777 & 0.864 \\
DeBERTa	&	0.754	&	0.739	 & \textbf{0.746} & \textbf{0.782} & \textbf{0.865}	\\
HateBert	&	0.755	&	0.719	&	0.737 & 0.777 & 0.857	\\
HateXplain	&	0.773	&	0.532	&	0.630 & 0.713 & 0.811	\\
\hline
    \end{tabular}
    \caption{Test classification results of binary incivility detection. ROC AUC and Macro-F1 summarize the results on the two classes.}
\label{tab:binary}
\end{footnotesize}
\end{table}

\begin{table}[t]
\begin{footnotesize}
    \centering
    \begin{tabular}{lccccc}
%        & \multicolumn{3}{c}{Uncivil}  \\
%\multicolumn{5}{l}
%{\textbf{Cross-dataset experiments:}} \\
Classifier	&	P	&	R	&	F1	& Mac.-F1 & AUC \\
\hline
{\it Theocharis}	&	0.73	&	0.61	&	0.665	&	\textbf{0.800} & -	\\
Ours &	0.542	&	0.847	&	0.661	&	0.782 & 0.848 \\
%Th & Ours	&	0.694	&	0.589	&	0.637 & 0.703 & 0.699	\\
\hline
{\it Davidson}  & -	&	-	&	-	&	0.802 & -	\\
Ours & 0.692	&	0.779	&	0.733	&	\textbf{0.850}	& 0.869\\
%D & Ours & 0.889 &	0.072	&	0.134 & 0.439 & 0.532\\
\hline
{\it Rheault (U)} &	-	&	-	&	-	&	0.738 & 0.763	\\
Ours &	0.549	&	0.841	&	0.665	&	\textbf{0.792} & \textbf{0.858}	\\
\hline
{\it Rheault (C)}  & -	&	-	&	-	&	0.763 & 0.766	\\
Ours & 	0.545	&	0.820	&	0.655	&	\textbf{0.801}	& \textbf{0.869}\\
\hline
    \end{tabular}
    \caption{Binary test classification results across datasets: previously reported within-dataset results (see Table~\ref{tab:stats}) vs. cross-dataset prediction performance using our RoBERTa finetuned model (`Ours').}
\label{tab:cross}
\end{footnotesize}
\end{table}

Table~\ref{tab:multi} details the results of the finetuned models on the test set in terms of ROC AUC, precision, recall and F1 with respect to each class, as well as Macro-F1 average over the two incivility types. We observe that all models achieve substantially lower performance in detecting intolerant as opposed to impolite speech, where the best F1 results obtained per these classes are 0.59 and 0.70, respectively. In line with the observed human agreement rates, this indicates that the automatic detection of political intolerance is a more challenging task.

The results of our binary classification experiments, considering political incivility as a unified concept, are given in Table~\ref{tab:binary}. As shown, coarse incivility prediction yields substantially higher results, reaching F1 of 0.78. In both setups, the best-performing classifiers are DeBERTa and RoBERTa. Henceforth, we consider RoBERTa as our classifier of choice, given its lower computational cost. 

To gauge the generality of our model and dataset, we also performed cross-dataset experiments. Table~\ref{tab:binary} includes the results of applying our binary model of political incivility detection  to other existing datasets (Table~\ref{tab:stats}), alongside the results previously reported per those datasets.\footnote{Our evaluation applies to randomly selected examples, except for a test set provided by~\cite{davidsonWS2020}.} As shown, our model gives best performance in almost all cases, \iffalse The dataset due to Theocharis {\it et al.} is an exception, where we achieve somewhat lower yet comparable F1 results (0.66 vs. 0.67).\fi showing high generalization across data distributions. \iffalse We further attempted to apply the models trained and distributed by~\citeauthor{theocharis20} and~\citeauthor{davidsonWS2020} to our test set. As detailed in Table~\ref{tab:binary}, this resulted in inferior F1 performance of 0.67 compared with our result of 0.75.\fi We consider this as indication for high diversity of our dataset.
 
\iffalse
The 2 latter dataset were labeled in figureEight and I don't have their exact test set so the results between their performance and ours are not on the same test set. In contrast, the first paper indeed publish the test set (Reddit posts) and it was labeled by three undergraduate students. Furthermore, their results may be lower depends on what we choose... specific details on the datasets can be found in "datasets overview.xlsx" attached 
ADD: Davidson--evaluated on their test set. Reault--evaluated on all of their datasets. Should check what they results pertain to. Theo--we evaluated on 20\%. They published on some other random 20\%. That is, all except first-not on the same test set. Theo: their data is more homogeneous (80\% of their dataset was tagged using Perspective).\fi

%\subsection{Additional analyses}

\paragraph{Impoliteness vs. intolerance detection.}

\begin{table}[t]
\begin{footnotesize}
\begin{tabularx}{0.48\textwidth}{X}
\hline 
{\it Impolite:} fuck, help, stupid, damn, obnoxious, fed, joke, ass, goddamn, shit, coward, crap, unreal, love, neoliberal, king, mentality, anarchist, fuel, publishing, bad, wow, back, bastard, communists, forgive, idiot, dumb \\
%, change, worst, terrible, broke, asshole, humiliating\\ 
\hline
{\it Intolerant:} republican(s),  democrat(s), leftists, GOP, democratic, catholics, speech, liberal, dem(s), socialist(s), conservatives, liberals, progressive(s), left, communist(s), party, right, racist, fascists, terrorists, nationalist(s) \\
%, constituents, marxist, whites, radical, destroyed, americans \\
%,  supremacist, congressional, fight, black(s), nationalist \\
\hline
\end{tabularx}
\caption{Salient unigrams associated with impolite and intolerant speech in our dataset (Shapley analysis)}
\label{tab:shap}
\end{footnotesize}
\end{table}

We applied Shapley analysis to our training set~\cite{shapNIPS17}\footnote{https://github.com/slundberg/shap} to identify unigrams predictive of political impoliteness and intolerance. As shown in Table~\ref{tab:shap}, impolite speech is characterised by derogatory words. Most of the listed words carry negative meaning in an unequivocal way, being offensive in any context, e.g., `stupid'. In contrast, we observe that the word types associated with political intolerance often refer to a political camp, e.g., `republicans', or `liberals'. Unlike slur words, the sentiment of such terms may depend on the context. In accordance, we found that impolite tweets were less susceptible to be classified as neutral compared with intolerant tweets (26.7\% vs. 44.0\%). This suggests that high-level semantic and contextual understanding is needed to detect intolerance. 

\iffalse
\subsection{CLS token+SV embeddings}
we use the output of the [CLS] token, a vector of length 768, from 12th transformer encoder concatenated with SocialVec vector embeddings of the tweet author and feed it as input to a fully connected neural network with one hidden layer. To classify each input sample a sigmoid activation function is employed to the hidden layer.
Getting SV embeddings requires a followees list for each tweet author. Using Twitter API we manage to extract the list for 2247 users (out of 3741) reflected in 9458 tweets (out of 13124). *there are users who do not exist anymore/are suspended...
see Table~\ref{tab:roberta+sv_statistics} and Table ~\ref{tab:roberta_sv_statistics_test} and images in 'roberta+sv' folder for precision/recall thresholds graphs.
\fi

%\paragraph{Error analysis.}

\iffalse
\begin{figure}[t]
\centering
\includegraphics[width=0.3\textwidth]{figures/confusion-matrix-new.png}
\caption{Confusion matrix at tweet level}
\label{fig:confusion}
\end{figure}
\fi

Examining the classification errors, we indeed observed cases for which the model missed the presence of intolerance due to its implied manifestation; e.g., ``you Republicans don't even know how to keep the electricity on!'', or, the sarcastic ``Don’t worry, the democrats are bringing in a billion illegal aliens to replace us with''. On the other hand, the model was sometimes misled by lexical cues, demonstrating the gap between lexical-level and semantic understanding; for instance, the tweet ``Yes I have hope for your country. There are enough people who are sick of this.'' was misclassified as impolite, possibly because of the idiom `sick of'. We further found some positive predictions of intolerance to be sensible while not being judged as such in the manual labeling process, demonstrating  the subtlety or subjectivity of this task; e.g., ``impeach Biden and his administration! Or charge them with treason''. Overall, these errors illustrate the challenge of semantic understanding for identifying political incivility. Ideally, relevant context information would be considered to improve the recognition of this phenomenon in general, and political intolerance in particular.

\paragraph{Impact of train set size.}

\begin{figure}[t]
\centering
\includegraphics[width=0.4\textwidth]{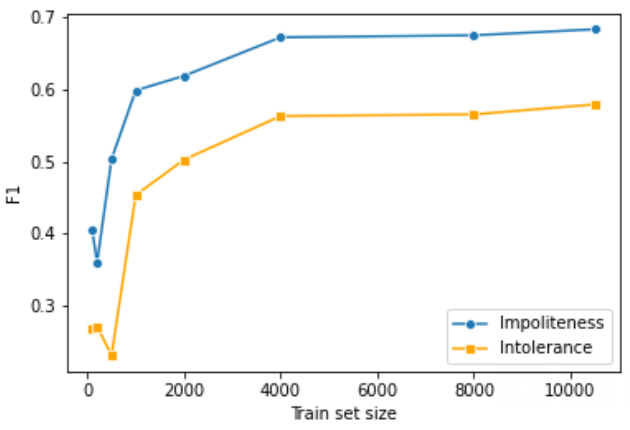}
\caption{Test F1 results on impoliteness and intolerance detection, varying the number of training examples.}
\label{fig:size}
\end{figure}

\iffalse Refresh: did the experiments on high-agreement tweets give higher results? (I believe they did, but only for the impoliteness class?) \fi

Figure~\ref{fig:size} shows test F1 results while finetuning our classifiers using increasing stratified subsets of the train set. It is shown that impoliteness detection dominates intolerance detection results using as few as 1,000 training examples. Again, we attribute this to the greater semantic complexity involved in political intolerance detection. Overall, the improvement in test performance subsides beyond $\sim$4K of labeled examples. Further improvements may be obtained by substantially extending the dataset via methods such as text generation~\cite{wullachEMNLP21} or back translation~\cite{ibrahim-etal-2020-alexu}. We leave this direction to future research.  

\section{From tweets to users: a large-scale evaluation}
\label{sec:large}

Next, we employ the learned models to identify, quantify and characterise political incivility at scale. In particular, we wish to explore whether certain users are more inclined to post politically uncivil content online, as well as to characterise such users. 

To address these questions, we considered a very large corpus of tweets, associated with author information. Concretely, we sampled user identifiers using Twitter API between July-November 2022, who were verified as U.S. residents based on the location attribute of their profiles. For each user, we retrieved their (up to 200) most recent tweets. Removing retweets, non-English tweets, tweets that only included URLs, and tweets posted by overly active accounts suspected as bots,\footnote{We removed accounts for which the tweet posting rate was higher than two standard deviations above the mean.} resulted in a corpus of 16.3M tweets authored by 373K users. Applying our classifier of political content detection, we obtained 2.57M political tweets authored by 230K distinct users, henceforth, {\it the corpus}. Overall, 17.6\% of the political tweets were identified as impolite, 13.3\% as intolerant, and 2.5\% as both categories, i.e., 28.4\% uncivil tweets overall. 

\subsection{Political incivility across subpopulations}

Using a corpus of political tweets that includes author information, one may investigate social factors that correlate with incivility. Below, we demonstrate this with respect to the social dimensions of political affiliation and state demographics.

\paragraph{Is incivility a matter of political affiliation?} To address this question, we gauged the prevalence of incivility among the two main political camps: {\it Democratic (liberal)} vs. {\it Republican (conservative)}. In this work, we opted for a simple and intuitive metric as a proxy of political affiliation. Considering the accounts of 30 popular news outlets scored by political bias~\cite{pew2020}, we identified users who followed two or more accounts included in this list, of homogeneous political orientation. Applying this criterion resulted in a sample of 54.5K users, out of which 83\% were assumed to be Democrats, and 17\% as Republicans. Our analysis showed minor differences between the two groups. The ratio of political impolite tweets was slightly higher within the Republican group (18.80\% vs. 18.52\%), whereas the ratio of politically intolerant tweets was higher among Democrats (9.06\% vs. 8.88\%), however neither of these differences was found to be statistically significant.

\paragraph{Do political incivility levels vary across states?}

To analyse and compare political incivility across U.S. states, we attended user accounts that specified state information (full state name, or its abbreviation) in the location meta-data field. Overall, 186K users in the corpus met this condition. The largest number of users were affiliated with the states of New-York (23K), California (16K) and Texas (14K). The states with the least number of users were North Dakota (265), Wyoming (315), South Dakota (426), and Alaska (579). The median number of tweets per state was 2,216, providing a sufficient sample size for statistical analysis. 

\begin{figure}[t]
\centering\includegraphics[width=.92\linewidth]{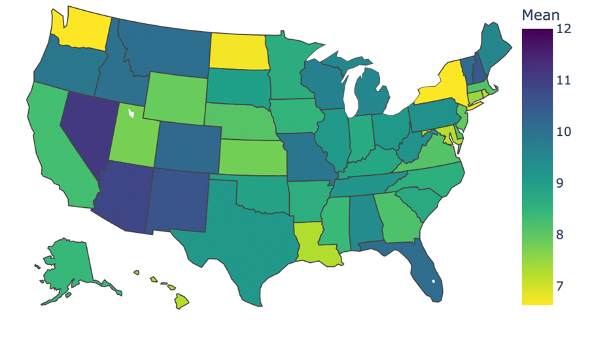}
\caption{Average detected user-level political intolerance ratio per state (ranging between 7-12\%).}
\label{fig:usamaps}
\end{figure}

For each state, we computed the average user-level proportion of impolite or intolerant tweets. Figure~\ref{fig:usamaps} presents a heat map showcasing the average intolerance ratio across states. Similar trends were observed for impoliteness. As shown, some states demonstrate low incivility rates (e.g., WA and NY) whereas other exhibit high incivility rates (e.g., AZ and FL). Presumably, in `battleground states', where the two camps are on par, there would be more hostility and toxicity in the political debate. To test this hypothesis, we compared the detected state-level average ratios of impolite and intolerant tweets against the differences between the percentage of votes for the Democratic and the Republican parties per state.\footnote{https://www.cookpolitical.com/2020-national-popular-vote-tracker} \iffalse Overall, there were 15 key battleground and 15 Non-Battleground states \textcolor{red}{explain}.\fi
Applying Spearman's correlation analysis confirmed our hypothesis, yielding correlation scores of -0.43 and -0.40, respectively, both found significant at p-value $<0.01$. In words, this result suggests that higher levels of political incivility in a particular state correspond to a closer contest between the two main political camps, manifested by a smaller difference in the vote percentage between the two parties.
\iffalse Conducting Spearman's correlation analysis, we found significant correlation between the average and median ratios of impolite tweets out of the political tweets and the voting gap between the camps.
Correlation scores were -0.43/-0.37, respectively, with p-value $>$ 0.01). The correlation between the average ratio of intolerant out of political tweets was also found to be significant (correlation score of -0.4, with p-value $>$ 0.01). (The median ratio of intolerant political tweets was nearly zero in all states.)
The correlation between the average and median ratios of uncivil out of political tweets was also found to be significant (correlation scores of -0.49, with p-value $>$ 0.001 / -0.44, with p-value $>$ 0.01).
In words, higher levels of political incivility in a particular state corresponds to a closer contest between the two main political camps, manifested by a smaller difference in the vote percentage between the two parties.\fi \iffalse States and their incivility statistics can be seen here: "Large Scale/242K users/incivility\_stats\_per\_state.csv".\fi

\subsection{Political incivility at user-level} 

Crucially, our results indicate that some users are more inclined to post uncivil content than others. The distribution of uncivil tweets in our corpus across users is highly skewed: as few as 7.3\% of the users authored 50\% of the uncivil posts in the corpus, and 20.6\% of the users authored 80\% of the uncivil posts. On the other hand, 43.7\% of the users authored no uncivil post.

\iffalse
\paragraph{Second-order analysis--paper\#2} We randomly sampled 1000 users collected between October 1st to October 25th 2022 and retrieved their followee’s tweets using Twitter API. To those 1000 users, there are 944060 followees overall (636647 are unique with 73773308 tweets). To deal with this enormous amount of tweets and given the fact that a user probably does not read all the tweets of the accounts that they follow, we filter tweets that were written in a period of two months before the collection date of the user. We were left with 7.996M tweets. Using the political classifier we found 513K political tweets and apply the incivility detection model. We aggregate for each one of the 1000 users the number of impoliteness tweets and intolerant tweets posted by his friends. attached correlation between the 1000 users and their friends, image in "Large Scale/1000 users" folder.
\fi

\begin{table}[t]
\centering
\begin{footnotesize}
\begin{tabular}{lcc}
& \% Impolite &	\% Intolerant \\
 \hline
\# Followers	&	-0.109	&	-0.038	\\
\# Friends	&	-0.017	&	0.058	\\
Tweets per day	&	0.068	&	0.091	\\
\% political tweets& 0.237 & 0.498 \\ 
\hline
\end{tabular}
\end{footnotesize}
\caption{Spearman's correlation: the ratio of impolite/intolerant political vs. other user-level metrics. All scores are significant (p-value$<0.001$).
}
\label{tab:users}
\end{table}

To further explore the distribution of incivility across users, we contrast the ratio of impolite and intolerant political tweets per user and other metrics of interest. As reported in Table~\ref{tab:users}, users who post intolerant and impolite political content tend to post more tweets per day. They also tend to have less followers--possibly, popular users refrain from controversial political language. Very high correlation was found between the ratio of intolerant and impolite tweets per user and the proportion of political tweets posted by them (Spearman's correlation scores of 0.50 and 0.24, respectively). \iffalse And, there is strong positive correlation between the prevalence of intolerance and impoliteness within one's political tweets (0.34).\fi That is, those users who discuss political topics more often, i.e., are more {\it politically engaged}, are more likely to use intolerant or impolite language.

\paragraph{A network perspective of user-level incivility.}
\label{sec:sample}

Next, we wished to explore whether social network information was indicative of one's tendency for using political uncivil language. 
In a controlled experiment, we sought differences between users who frequently post politically uncivil content and users who rarely do so. \iffalse More precisely, while we acknowledge that there are contextual factors that may evoke incivility, we examine whether there are user-level network characteristics that are indicative of the inclination of turning to uncivil language in discussing politics.\fi 
In our analysis, we considered a random sample of 1,000 user accounts from our corpus for which we identified a high ratio of incivility (above 50\%) within their political posts. For each selected user, we identified a counter example--another user with a similar ratio of political tweets, and no indication of incivility. As a result, the proportion of political tweets per user in the two groups is similar (roughly 37\%). But, while the ratio of incivility within the political tweets in the first group is high (roughly 34\% impolite, 39\% intolerant, and 66\% uncivil overall), the prevalence of political incivility within the control group is practically zero, by design. 

The users in both groups follow about the same number of accounts on average. Yet, we found differences in the types of accounts that each group tends to follow. To identify such accounts, we computed pointwise mutual information (PMI) scores as follows: $log {Pr(s_i,a_j) \over Pr(s_i)*Pr(a_j)}$, where $a_j$ denotes some account followed, $Pr(s_i,a_j)$ is the joint probability that users of group $s_i$ follow account $a_j$, and $Pr(a_j)$ is the probability that any user, of either group, follow that account. \iffalse A positive PMI score indicates that a user in the uncivil group is more likely to follow account $e$ compared with a random user in our sample of political users, and vice versa.\fi High PMI scores indicate on strong correlation, whereas low (near zero) scores correspond to independent events. 

%We considered about 4.8K accounts followed by more than 1\% (20) of the users in our sample. 
Manually examining the accounts that characterize the users who post uncivil content, we found that many of them deliver a political message in their account description, e.g.: ``celebrating Trump-free gov't'', ``\#ResistFascism'',``\#nonazis'', ``NoGoZone for Democrats, Socialist, Globalist, and Godless AntiAmericans.'', or ``\#TrumpWonBidenCheated''. In contrast, the counter `political, yet civil' group of users was found to distinctively follow political organizations, charitable foundations, as well as economical, scientific, and technological news sources and columnists. Overall, these exploratory results suggest the network profile of users encodes meaningful social context information that correlates with political incivility.

\section{User-informed incivility detection}

Having established that some users post political uncivil content more than others, and that there are meaningful network cues that characterise those users, we argue and show that the joint modeling of tweets and their authors can improve the performance of automated political incivility detection.

\iffalse This result is important in that not only can traits such as gender or education level can be predicted based on public user information in the social network, but also the inclination of that individual to use uncivil rhetoric in discussing political issues. Hence. we believe that it can serve as relevant background information in applying automated methods for decoding the intent that underlies text posted by them in general, and for detecting expressions of political incivility in particular. \fi

\subsection{Approach}

\begin{table*}[t]
\begin{footnotesize}
    \centering
    \begin{tabular}{l|ccc|ccc|ccc}
    & \multicolumn{3}{c}{Impolite} & \multicolumn{3}{c}{Intolerant} & \multicolumn{3}{c}{Uncivil} \\
Evidence &	P &	R &	F1	& P & R	&	F1 & P & R & F1 \\
    \hline
        tweet  & 0.617 & 0.818 & 0.704 & 0.403 & 0.703 & 0.512 & 0.774 & 0.583 & 0.665   \\ 
        tweet \& user: all (encoding)   & 0.628 & 0.800 & 0.704 & 0.403 & 0.710 & 0.514 & \textbf{0.778} & 0.647 & 0.707 \\
        tweet \& user: list-based (encoding) & 0.630 & 0.800 & 0.705 & 0.396 & 0.712 & 0.509 & 0.737 & 0.684 & 0.709 \\
        tweet \& user: sample-based (encoding) & 0.647 & 0.798 & 0.714 & 0.395 & 0.710 & 0.508 & 0.716 & \textbf{0.718} & 0.717 \\ 
        % Tweet + High-PMI SV($>$ 0.1) & 0.658 & 0.791 & 0.718 & 0.400 & 0.740 & 0.520 & 0.619 \\ 
        % Tweet + High-PMI SV($>$ 1) & 0.654 & 0.787 & 0.714 & 0.419 & 0.706 & 0.526 & 0.620 \\
        tweet \& user: high PMI (0.5) (encoding) & 0.608 & \textbf{0.831} & 0.702 & 0.407 & 0.721 & 0.520 & 0.778 & 0.615 & 0.687 \\ 
        tweet \& user: high PMI (1.0) (encoding) & \textbf{0.665} & 0.784 & \textbf{0.720} & 0.409 & \textbf{0.725} & \textbf{0.523} & 0.754 & 0.693 & \textbf{0.722} \\
        tweet \& user: high PMI (1.0) (sparse) & 0.657 & 0.771 & 0.710 & \textbf{0.420} & 0.639 & 0.507 & 0.730 & 0.693 & 0.713 \\
                \hline
    \end{tabular}
\caption{User-informed political incivility detection test results}
\label{tab:contextual}
\end{footnotesize}
\end{table*}

\paragraph{User encoding.} One may represent users in terms of relevant accounts that they  follow using sparse binary indications~\cite{lynn19}. Here, we rather exploit account embeddings, learned from a large sample of the Twitter network for this purpose~\cite{lotanPLOS23}. Given a sample of 2M U.S. Twitter users and the accounts that they follow, the embeddings of 200K popular Twitter accounts were learned, such that accounts which users tend to co-follow are placed close to each other in the embedding space. Consequently, the embeddings encode social and topical similarities.  We project individual users onto the social embedding space by averaging the embeddings of accounts of interest that they follow.  

\iffalse Previously, researchers represented users using binary indications of whether they followed specific accounts. This type of evidence yielded higher performance on the task of stance detection compared with other meta-data and content information associated with those users. This representation form is sparse however. \fi

\paragraph{A unified classification approach.} The semantic encoding of a given tweet and the  social encoding of the tweet author are incompatible, yet we wish to combine them in performing political incivility detection. Our proposed approach consists of the following principles. We obtain the encoding of a given tweet output by the finetuned transformer-based RoBERTa model. We then concatenate this content encoding with the respective social user encoding of the tweet author. This multi-facet evidence is served into a dedicated multi-layer neural network, which we train, tune and test using our training, validation and test examples. 

\subsection{Experiments}

\iffalse
\begin{table*}[h]
\begin{footnotesize}
    \centering
    \begin{tabular}{llccccc}
%        & \multicolumn{3}{c}{Uncivil}  \\
\multicolumn{2}{l}{Classifier}	&	P	&	R	&	F1	& Mac.-F1  \\
\hline
\multicolumn{2}{l}{Tweet encoding}	&	0.774	&	0.583	&	0.665	& 0.699  \\
\multicolumn{2}{l}{Tweet + SV(all)}	&	0.778	&	0.647	&	0.707	& 0.726  \\
\multicolumn{2}{l}{Tweet + SV(popular, $>$ 20)}	&	0.716	&	0.718	 & 0.717 & 0.713  \\
% \multicolumn{2}{l}{Tweet + High-PMI SV($>$ 0.1)}	&	0.776	&	0.640	&	0.701 & 0.722  	\\
% \multicolumn{2}{l}{Tweet + High-PMI SV($>$ 1)}	&	0.757	&	0.670	&	0.711 & 0.723  	\\
\multicolumn{2}{l}{Tweet + High-PMI SV($<$ -0.5,$>$ 0.5)}	&	0.778	&	0.615	&	0.687 & 0.713  	\\
\multicolumn{2}{l}{Tweet + High-PMI SV($<$ -1,$>$ 1)}	&	0.754	&	0.693	&	0.722 & 0.729  	\\
\multicolumn{2}{l}{Tweet + Sparse vector}	&	0.73	&	0.693	&	0.713 & 0.717 	\\
\multicolumn{2}{l}{Tweet + SV (political)}	&	0.737  	&	0.684	&	0.709 & 0.715	\\
\hline
    \end{tabular}
    \caption{Binary test classification results: Precision, recall and F1 apply to the incivility class. Macro-F1 summarizes the results on the two classes. - updated}
% \label{tab:binary}
\end{footnotesize}
\end{table*}
\fi

To perform user-informed tweet classification, we obtained the list of accounts followed by each user in our dataset using Twitter API. At the time of network data collection, we were able to retrieve relevant information for 2,247 (out of 3,741) distinct users; some users may have been suspended, or quit the social network. This yielded a smaller dataset of 9,458 labeled tweets with available author network information. The distribution of labels remained similar to the original dataset (59.1\% neutral, 26.0\% impolite, 16.7\% intolerant, and 2.0\% labeled as both). In conducting user-informed tweet classification, we split this dataset into class-stratified sets, and further verified that there was no overlap between the authors of tweets in the test  set and the examples used to train and tune the models.  

In the experiments, we finetuned RoBERTa and extracted a 768-dimension CLS vector from the classifier as the tweet encoding. We obtained 100-dimension social embeddings of relevant accounts that each user followed, aggregating them into an averaged user encoding.\footnote{https://github.com/nirlotan/SocialVec} The concatenated tweet and author embeddings were fed into a fully connected neural network with a Sigmoid output unit. We learned models for detecting impoliteness and intolerance, as well as a binary notion of political incivility. In learning, we minimized a binary cross-entropy loss function, while tuning the hyper-parameters of the neural network, including the learning rate, optimizer, the number of hidden layers and their size. \iffalse between 1e-5 to 1e-1, the number of hidden layers ranging from 1 to 3, and each layer's size between 4 to 18. Additionally, I tested multiple optimizers such as adam, adamW, SGD, and RMSprop using the Optuna package. \fi Considering the reduced dataset size, we performed tuning using cross-validation, and trained the final models using the full train set.

\subsection{Results}

Table~\ref{tab:contextual} details our results on multidimensional and binary political incivility detection evaluated on the test set. \iffalse We performed also cross-validation (CV) experiments, which showed consistent results and trends with the outcomes reported in this table. The results of the CV experiments are included in Appendix x.\fi The table includes the results using the tweet alone as baseline. We note that the reported performance is overall lower compared with our previous experiments--we attribute this to the reduced dataset size. 

As detailed in the table, we report the results of user-informed incivility detection using several different user representation schemes. Concretely, we attempted representing the users in terms of all of the accounts that they follow (`all'), or in terms of the following account subsets of interest:

--Accounts that are known to be politically biased according to external lists; see Sec.~\ref{par:political} (`List-based').

--Popular accounts followed by 1\% or more of the users in our sample of 2K highly political users, who exhibit either high or low incivility; see Sec.~\ref{sec:sample} (`Sample-based').

--We further narrowed the previous sample-based subset to those accounts that are distinctive of high or low incivility, with absolute PMI scores greater than 0.5 or 1.0 (`Sample-PMI')

As shown, major improvement were achieved using all methods on binary incivility detection, reaching an impressive improvement of up to 5.7 absolute points in F1. Lower, yet substantial, improvements were also achieved on impoliteness and intolerance detection, reaching gains of up to 1.6 and 1.1 absolute points in F1, respectively. In all cases, the best results were obtained by focusing on network information that was found distinctive of political incivility in our analysis (sample-based, PMI 1.0). As expected, representing the user's network information in terms of social embeddings is beneficial compared with a sparse representation of the same set of accounts (`sparse'). Overall, we find these results to be highly encouraging, indicating that the social modeling of users provides meaningful contextual evidence that improves the decoding of the texts that they author.

\section{Conclusion}

This work framed political incivility detection as a multidimensional classification task, distinguishing between impolite and intolerant political discourse. We collected a large dataset of multidimensional political incivility, annotated via crowd sourcing, which we believe is diverse and representative of the challenges that automated political incivility detection must address. In particular, we observed high lexical ambiguity and a need for incorporating semantic and social cues in decoding political intolerance. In a large-scale study, we showcased various social factors correlated with political incivility that apply to subpopulations and individual users. Last, we leveraged relevant social network information, presenting substantial improvements in incivility detection by augmenting the textual evidence with social context information about the text author. We believe that this research direction holds promise for social text processing in general. 

\section{Limitations}

While we targeted the detection of political intolerance as a broad concept, we observed that the tweets annotated as intolerant in our dataset often aim to undermine or silence specific partisan and political groups (e.g., `republicans', `democrats', or `liberals'). Other flavors of political intolerance, including expressions toward immigrants, ethnic minorities, or other social groups, may be underrepresented in the dataset. It is possible that our political content classifier contributed to this bias, or that  political intolerance in its bipartisan context is inherently more prevalent in Twitter. In addition, our study applies to political incivility in the U.S., focusing on the Twitter network. While we believe that our model and insights are general to a large extent, they may be limited geographically, temporally, and across social media platforms.

\paragraph{Ethical considerations.} Despite analyzing incivility at user-level, we emphasize that political incivility is common, context-dependent, and should not be considered as a personal characteristic. This research was approved by our institutional review board. We release our code and dataset, adhering to Twitter terms, to promote future related research.

\newpage

\bibliography{final}
\bibliographystyle{acl_natbib}

\end{document}